\documentclass[11pt]{article}

\usepackage[final]{acl}
\usepackage{placeins}

\usepackage{times}
\usepackage{latexsym}

\usepackage[T1]{fontenc}

\usepackage[utf8]{inputenc}

\usepackage{microtype}

\usepackage{inconsolata}

\usepackage{amsmath}
\usepackage{amssymb}
\usepackage{graphicx}

\usepackage[table, dvipsnames]{xcolor}

\usepackage{booktabs}

\definecolor{lightblue}{RGB}{220,240,255}
\definecolor{lightgreen}{RGB}{220,235,220}
\definecolor{lightyellow}{RGB}{255,250,220}

\definecolor{softpink}{RGB}{250,220,240}     
\definecolor{softblue}{RGB}{210,230,250}     
\definecolor{softorange}{RGB}{255,230,200}   
\definecolor{softgreen}{RGB}{220,245,220}    
\definecolor{softgray}{RGB}{240,240,240} 

\title{Visual Search Augmented Chain-of-Thought Reasoning for Attribute Value Extraction from Product Videos}

\author{
\bf Tong Wu$^1$, Ming Cheng$^1$, Jiazhen Hu$^1$, Jiaying Gong$^2$\thanks{This work was completed prior to joining Amazon, and does not relate to the author's position at Amazon.}, Hoda Eldardiry$^1$ \\
$^1$Virginia Tech, $^2$Amazon \\
\texttt{\{tongw,ming98,hjiazhen,hdardiry\}@vt.edu, gojiayin@amazon.com}
}

\begin{document}
\maketitle
\begin{abstract}
Existing approaches to visual attribute value extraction (AVE) primarily rely on static product images, failing to capture temporal cues, multi-angle views and fine-grained visual details.
Directly applying video vision-language models (VLMs) to product AVE results in limited performance due to the lack of domain knowledge, and fine-tuning them requires extensive high-quality data and substantial computational resources.
Thus, we propose visual search augmented chain-of-thought reasoning (ViS-CoT), a training-free, plug-and-play pipeline that can be easily applied to any open-source video VLM for video-to-text AVE in e-Commerce.
Specifically, ViS-CoT employs visual clustering to identify representative frames, followed by visual search to retrieve semantically similar product knowledge that can enrich attribute cues.
Next, an interleaved CoT reasoning module iteratively refines reasoning through visually-aligned auxiliary texts derived from captioning and automatic speech recognition. 
Finally, the integrated information guides the model toward accurate and fine-grained attribute predictions. Extensive experiments across 14 product categories on the VideoAVE dataset show that ViS-CoT consistently enhances multiple state-of-the-art video VLMs, achieving an average improvement of \textbf{17.91} percentage points in micro-F1.
\end{abstract}

\section{Introduction}
Visual attribute value extraction (AVE) is a core problem in computer vision and e-Commerce that seeks to automatically identify and structure fine-grained product information such as brand, material, types, and item forms from multimodal content. Reliable extraction of such attributes is crucial not only for improving product retrieval, recommendation, and filtering, but also for enabling downstream tasks such as large-scale catalog management and standardization across sellers and marketplaces~\cite{gong-etal-2025-mice}.

\begin{figure}[htp] 
 \center{\includegraphics[height=3.5cm,width=7.5cm]{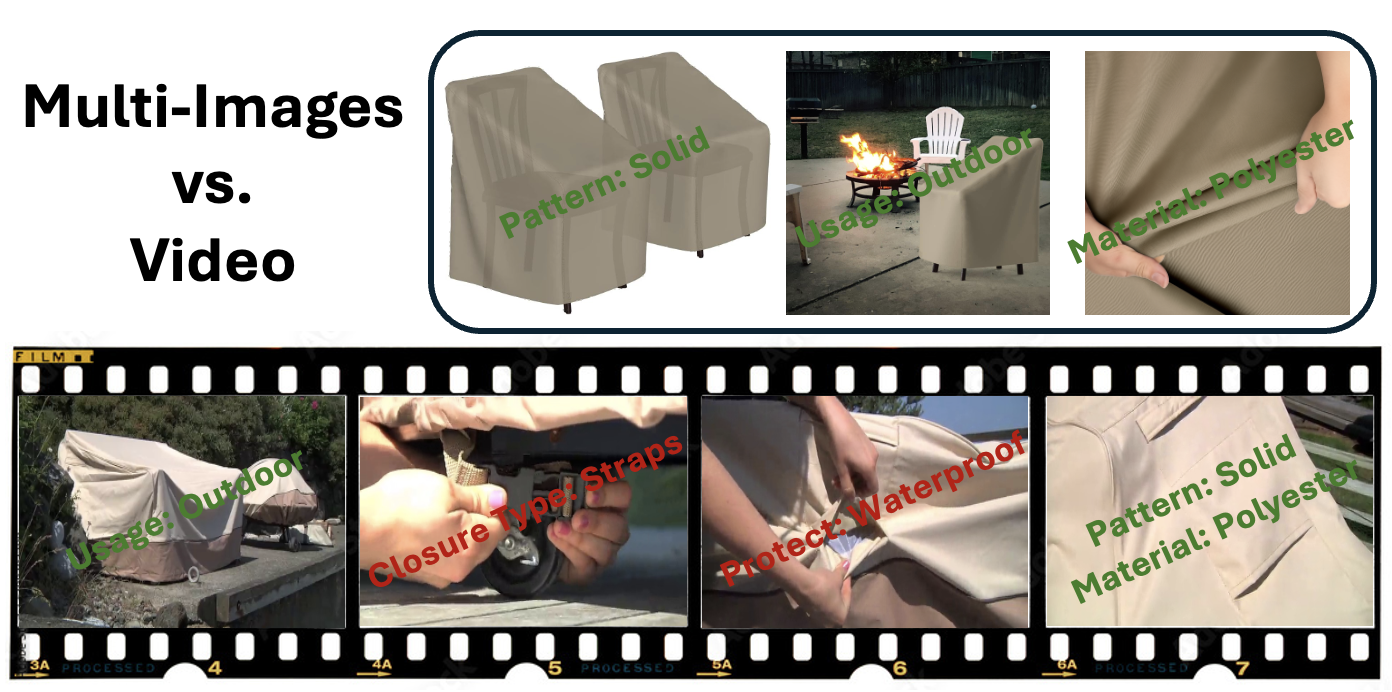}}
 \caption{\label{fig:intro} Comparison of product attribute value generation from multi-images vs. from videos.}
 \vspace{-4mm}
 \end{figure}

Existing research on product visual AVE has primarily relied on static product image inputs~\cite{gong-etal-2025-mice, gong-etal-2025-visual, zou-etal-2024-eiven, zou-etal-2024-implicitave, khandelwal-etal-2023-large}.
However, many attributes cannot be reliably inferred from the product images alone, because the static nature of single or even multiple product images often fails to capture fine-grained details, temporal cues, and multi-angle views.
Figure~\ref{fig:intro} shows that relying solely on multi-image sets may overlook fine-grained usage attributes. 
While static images display general characteristics, they often fail to capture key functional details such as closure type and protection type, which are more effectively revealed through dynamic video frames.
In contrast, product videos can provide richer visual information as they cover all sides of an item through continuous rotation, revealing subtle product properties (i.e. color, material, etc.) under changing environments (i.e. lighting condition, usage scenarios, etc.).
These dynamic views help reduce ambiguity, lower the chance of missing attributes, and improve the alignment with how consumers perceive products in real e-Commerce platforms, leading to more accurate and complete attribute value extraction.

Recent work~\cite{10.1145/3746252.3761621} introduced video-to-text AVE benchmarks to address the limitations of product image inputs. However, current video VLMs typically rely on uniform frame sampling, which often captures product-irrelevant, noisy or redundant content which results in hallucinations.
While advances in general video understanding (i.e. adaptive trees~\cite{wang2025videotree}, differential sliding-window captioning~\cite{chen2024sharegpt4video}, etc.) have improved semantic modeling of video dynamics, these methods differ fundamentally from product AVE.
General video understanding emphasizes what happens in the video (events, narratives, storytelling), whereas product video understanding focuses on what the item is and its properties, leveraging motion primarily to reveal attributes rather than to construct a narrative.

To address the limitations, we propose \textbf{ViS-CoT}, a visual search augmented chain-of-thought framework for product video understanding. The key idea is to combine retrieval-based grounding with staged multimodal reasoning.
Unlike general video understanding for storytelling, product video understanding aims at extracting structured, fine-grained product attributes by solving visual ambiguity between attributes. ViS-CoT consists of four stages:
(1) \textbf{key frame extraction}, where visual clustering selects representative frames from the video;
(2) \textbf{visual search}, which retrieves similar products to provide attribute priors;
(3) \textbf{interleaved CoT reasoning}, which iteratively refines attribute hypotheses using captions and ASR transcripts; and
(4) \textbf{integration and generation}, where multimodal evidence is combined to produce final attribute predictions. We evaluate ViS-CoT across 14 categories on a video-to-text e-Commerce dataset VideoAVE~\cite{10.1145/3746252.3761621}.
By applying ViS-CoT to five SOTA open-source video VLMs, we achieve an average improvement of \textbf{17.91} percentage points in micro-F1 on VideoAVE, while achieving competitive performance compared with fine-tuned video VLMs.

Our main contributions:
\begin{itemize}
    \item We introduce \textbf{ViS-CoT}, the first plug-and-play visual search-augmented interleaved CoT pipeline designed to enhance any video VLM.  ViS-CoT is training-free and broadly applicable, offering a resource-efficient solution for video-to-text AVE in e-Commerce.
    \item We observed a key technical insight: the \textbf{two-stage interleaved design} (vision-only hypothesis $\rightarrow$ retrieval-grounded refinement) is more effective than using visual search or CoT individually, improving attribute prediction by grounding reasoning in retrieved product knowledge.
    \item We conduct extensive experiments across 14 product categories using five state-of-the-art video VLMs, demonstrating significant performance improvements and validating the contribution of each component.
\end{itemize}

\section{Related Work}
\label{sec:RW}
\paragraph{Attribute Value Extraction}
Early research on AVE has mainly focused on unimodal approaches, where product attributes are derived solely from textual inputs such as titles or descriptions~\cite{10.1145/3589334.3645649, fang2024llm, li2024ecomgpt, 10.1145/3583780.3615142, deng-etal-2023-product, shinzato-etal-2023-unified, yang2023mixpave, blume2023generative, brinkmann2025self, su2025taclr}.
However, by relying only on text, these models often fail to capture the rich visual information and multimodal correlations in product images.
Subsequent work has shifted toward vision language models (VLMs)~\cite{Qwen2.5-VL, grattafiori2024llama, chen2024internvl, hong2025glm, comanici2025gemini, li2024llava, achiam2023gpt, abdin2024phi}, to capture the rich visual information and multimodal correlations in product images~\cite{gong-etal-2025-mice, trabelsi2025matters, gong-etal-2025-visual, hu2025hypergraph, hongwimol2025gavel, zou-etal-2024-implicitave, zou-etal-2024-eiven, khandelwal-etal-2023-large, zhang-etal-2023-pay}.
While VLMs enable the inference of some implicit attributes that are not explicitly stated in product descriptions (i.e., material, pattern, etc.), they are still limited by the static nature of single-frame inputs, which fail to capture temporal or multi-angle product views.
To enhance the understanding of both spatial and temporal cues inherent in video data, we further explore the video contents including audio transcripts, object locations, and scene-level captions on VideoAVE~\cite{10.1145/3746252.3761621}.

\paragraph{Video Understanding}
Though recent VLMs have extended their capabilities to video data~\cite{wang2024internvideo2, Qwen2.5-VL, lin-etal-2024-video, damonlpsg2023videollama}, these video VLMs still struggle to achieve fine-grained task-specific understanding~\cite{tang2025video}.
Task-specific video understanding models could be roughly categorized into classification~\cite{rehman2021deep, tan2024overlooked, han2024video}, captioning~\cite{qasim2025dense, lian2025describe, chen2024sharegpt4video, islam2024video, xu2024retrieval, kim2024you}, and localization approaches~\cite{khosla2025relocate, hu2025exploiting, feng2025object, pei2025object, kumar2024qdetrv, wang2024omnivid} in diverse areas such as automatic driving~\cite{fang2024abductive, wang2024drivedreamer}, robotics~\cite{mccarthy2025towards, sermanet2024robovqa}, science and healthcare~\cite{zhao2025mmvu, shikino2025understanding}, etc.
Building on recent advances in retrieval-augmented video comprehension~\cite{jeong-etal-2025-videorag, zheng2025retrieval, VideoRAG, mao2025multi, arefeen2024irag, arefeen2024vita, luo2026video}, which incorporate contextually relevant knowledge into the interpretation of video content, we leverage visual search to enhance interleaved chain-of-thought (CoT) reasoning for AVE in e-Commerce.

\begin{figure*}[t!] 
 \center{\includegraphics[height=8.5cm,width=\textwidth]{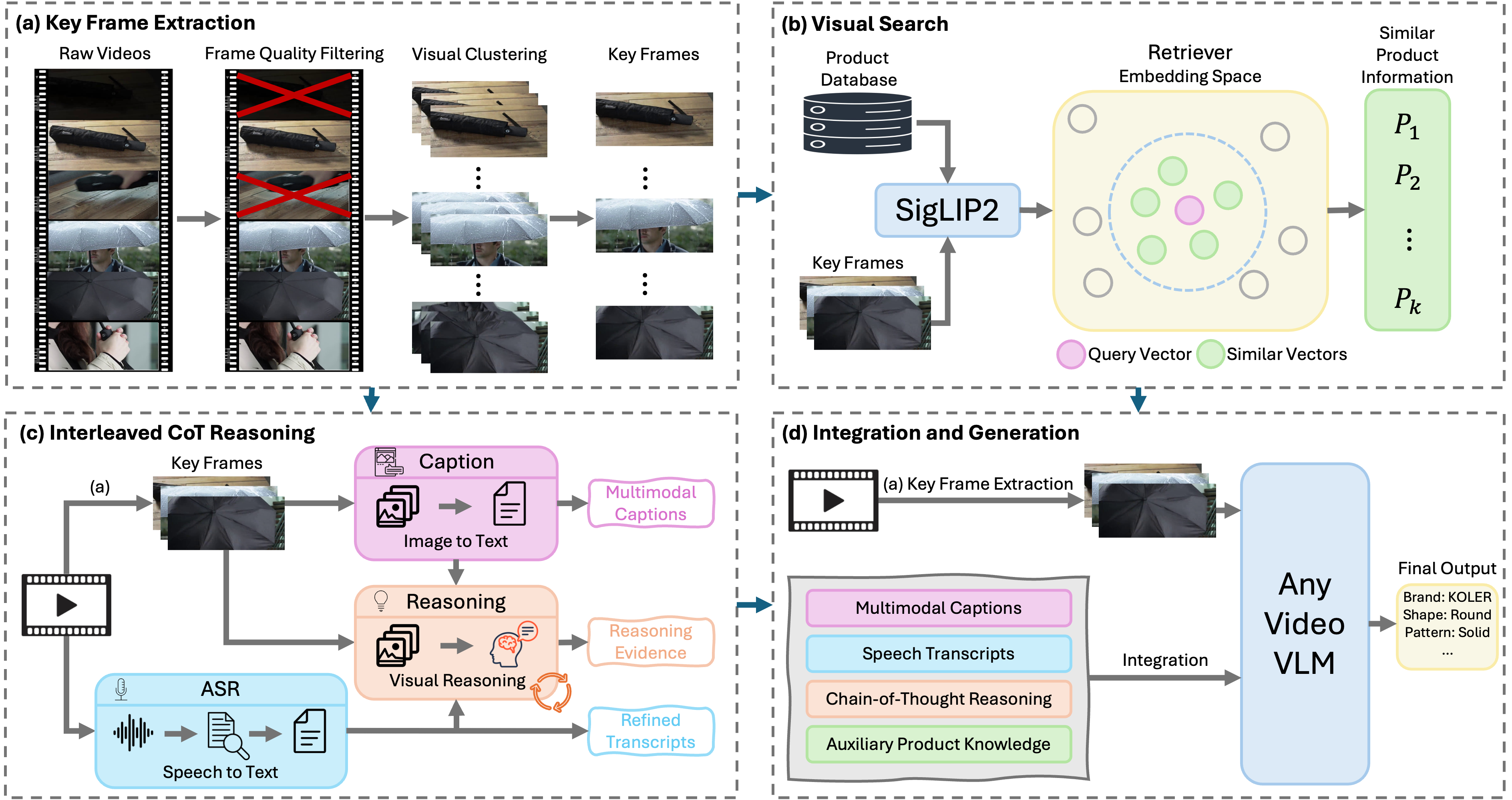}}
 \caption{\label{fig:framework} The framework of ViS-CoT, which includes four stages. In stage (a), key frames are extracted from the raw product video through filtering and clustering. In stages (b) and (c), video and the key frames are processed to generate four types of textual information: (i) similar product information via visual search, (ii) image captions, (iii) refined transcripts from ASR, and (iv) interleaved CoT reasoning. In stage (d), all auxiliary textual information from stages (b) and (c) are integrated with the key frames to produce the final answer.}
 \vspace{-4mm}
 \end{figure*}

\paragraph{Chain-of-Thought Reasoning}
Chain-of-thought (CoT) reasoning enables models to decompose a problem into intermediate steps, forming a structured reasoning path to facilitate multi-step inference~\cite{wang2025multimodal}.
With the rapid development of VLMs, the integration of CoT reasoning into multimodal domain (MCoT) has gained increasing attention~\cite{zhou-etal-2025-miceval, chen-etal-2024-m3cot, tan2024boosting, zhang2023multimodal, zheng2023ddcot}.
MCoT strategies typically fall into three categories: prompt-based methods~\cite{fei2024video, wu2024dettoolchain, gao2024cantor, 10.5555/3737916.3742339, wang2024videoagent}, plan-based methods~\cite{Zhang_2025_CVPR, yang-etal-2024-soft, zheng2024picture}, and learning-based methods~\cite{10656469, wang2024t, tan2024boosting, zhong2024let}.
These models often adopt structured pipeline and learn reasoning patterns from training data to guide multimodal reasoning.
To enable more comprehensive reasoning from multimodal inputs, recent studies explore the integration of expert tools (i.e. LLM expertise, auxiliary visualization, etc.)~\cite{10.5555/3737916.3742339, wu2024dettoolchain, gao2024cantor, li-etal-2024-enhancing-advanced}, and internal or external knowledge (i.e. knowledge graphs, object localization, etc.)~\cite{mitra2024compositional, mondal2024kam, khaliq-etal-2024-ragar, jia2024dcot}.
However, these MCoT approaches are limited by the fixed general-purpose pipelines designed for visual question answering, which causes inefficiencies and the inclusion of redundant or noisy information for e-Commerce product video understanding.

\section{Methodology}

We propose ViS-CoT, a training-free visual search augmented reasoning pipeline for large VLMs, which can be integrated into any video-language models. As shown in Figure~\ref{fig:framework}, ViS-CoT consists of four major components: (1) key frame extraction, (2) visual search, (3) interleaved CoT reasoning, and (4) integration and generation.

\subsection{Problem Formulation}
We consider the task of multimodal product AVE, where the raw input is the product video $\mathcal{V}_{p}$.
Following the two settings described in~\cite{10.1145/3746252.3761621}, we consider both the attribute-conditioned and the general formulation.
In the attribute-conditioned setting, the model is provided with a predefined list of attributes and is tasked with extracting their corresponding values from the video: 
\begin{equation}
f_{\theta}(\mathcal{V}_{p}, \left\{a_{1}, a_{2}, \cdots, a_{m} \right\})  \to  \left\{v_{1}, v_{2}, \cdots, v_{m} \right\}
\end{equation}
where $V_{p}$ is the product video, $a_{i}$ is the pre-defined attribute, and $v_{i}$ is the extracted value, $m$ denotes the total number of attributes. 
The general setting is formulated as an open attribute-value pair extraction problem, where no attributes are specified in advance.
The model infers both relevant attributes and their associated values:
\begin{equation}
f_{\theta}(\mathcal{V}_{p}) = \left\{(a_{1}, v_{1}), (a_{2}, v_{2}), \cdots, (a_{m}, v_{m}) \right\}
\end{equation}
The raw input video $\mathcal{V}_{p}$ is then refined and enriched with auxiliary information, such as transcripts, metadata retrieved through visual search, and other contextual cues, as described in Sec.~\ref{sec:key}, Sec.~\ref{sec:vs}, and Sec.~\ref{sec:CoT}.

\subsection{Key Frame Extraction}\label{sec:key}
\paragraph{Frame Quality Filtering.}
The first step of our framework automatically filters low-quality frames from product videos.
Specifically, we remove frames that are blurry, overly black or white, or exhibit low contrast. The cleaned set of video frames for a product $F^{c}_{p}$ is defined as:
\begin{equation}
\begin{aligned}
F^{c}_{p} = \{ I \in F_p \;|\; & v_{\mathrm{Lap}}(I) \ge \tau_{\mathrm{blur}} \\
& \wedge \;\tau_{\mathrm{black}} \le \mu(I) \le \tau_{\mathrm{white}} \\
& \wedge \;\sigma(I) \ge \tau_{\mathrm{contrast}} \}
\end{aligned}
\end{equation}
where $v_{\mathrm{Lap}}(I)$ is the Laplacian variance measuring the sharpness of the video frame image $I$, $\mu(I)$ is the mean intensity capturing brightness, and $\sigma(I)$ represents contrast. $\tau_{\mathrm{blur}}$, $\tau_{\mathrm{white}}$, $\tau_{\mathrm{black}}$, and $\tau_{\mathrm{contrast}}$ are thresholds for these attributes. 
This step ensures that only visually informative frames are retained, which is critical for preserving the quality and usability of the video data for downstream tasks.

\paragraph{Visual Clustering.}
After filtering out uninformative frames of a product video, there are still large quantities of redundant or irrelevant content.
To reduce redundancy, we introduce a visual clustering step that groups the cleaned video frames $F^{c}_{p}$ based on semantic similarity. This enables the model to focus on representative product-related frames from each cluster while discarding repetitive or irrelevant content. 
Given a set of cleaned video frames $F^{c}_{p}= \left\{f^{c}_{1}, f^{c}_{2}, \cdots, f^{c}_{n} \right\}$, we first extract visual features using a pre-trained encoder $f_{i}=E(f^{c}_{i})$. 
Next, we use K-Means clustering to group into $m$ clusters:
\begin{equation}
\min_{\{C_j\}_{j=1}^m} \sum_{j=1}^{m} \sum_{f_i \in C_j} \lVert f_i - \mu_j \rVert^2
\end{equation}
where $\mu_j$ is the centroid of cluster $C_j$.
From each cluster, we select the frame closest to its centroid as the key frame $f^{key}_{i}$.
This clustering process partitions the original $n$ frames into $m$ clusters ($m \ll n$), resulting in a non-redundant and diverse set of frames $F^{key}_{p}$ that efficiently and effectively represent the product.

\subsection{Visual Search}~\label{sec:vs}
We first construct a database from the product images available in the training data.
To enable efficient and accurate retrieval during the visual search phase, each training image is encoded into a feature vector $\mathbf{x}_i \in \mathbb{R}^d$ using the visual encoder of SigLIP2~\cite{tschannen2025siglip2multilingualvisionlanguage}, which is stored alongside its product identifier.  
During inference, query key video frames $F^{key}_{p}$  are encoded by the same pretrained SigLIP2 to obtain query embeddings $ \left\{q_{1}, \cdots, q_{m} \right\}$.
The cosine similarity between query and database embeddings is then computed, and to ensure robustness across multiple query frames, the similarity scores are averaged. The Top-$k$ nearest neighbors are then retrieved as:

{\tiny
\begin{equation}
\{P_{1},\cdots,P_{k}\} =
\operatorname*{arg\,top}_{k}\big( \{ \frac{1}{m} \sum_{j=1}^{m} 
    \frac{\mathbf{q}_j^\top \mathbf{x}_i}{\|\mathbf{q}_j\| \, \|\mathbf{x}_i\|} \mid i = 1, \dots,N \} \big)
\end{equation}}
where $P_{i}$ denote the retrieved top-$k$ products and their associated attribute values.
Finally, the top-$k$ retrieved products form the candidate aspect set for the query product.

\subsection{Interleaved Chain-of-Thought Reasoning}~\label{sec:CoT}
We generate the auxiliary texts from the video, followed by an interleaved chain-of-thought reasoning step, which iteratively refines understanding by alternating between the generated reasoning and the auxiliary context evidence.

\paragraph{Auxiliary Information Generation.}
To provide richer semantics, context, and evidence for the following reasoning step, we generate two different forms of auxiliary information that are fed into the interleaved CoT reasoning step.
(1) \textit{Captions $D_{p}$}. To capture semantic details from the key frames $F_{p}^{key}$ in Sec.~\ref{sec:key}, we leverage a  VLM-based captioner to convert visual content into textual descriptions: $D_{p}=Caption(F_{p}^{key})$. These captions serve as detailed descriptions of the semantic content from the corresponding key frames.
(2) \textit{Transcripts $T_{asr}$}. Audio signals provide complementary information for video understanding, often providing additional context that may not be available through visual cues alone. For each product video $\mathcal{V}_{p}$, we first extract the raw audio $U$ and transcribe it into text by an Automatic Speech Recognition (ASR) module, specifically Whisper~\cite{radford2023robust}. To further enhance conciseness and informativeness, we apply a VLM-based summarizer to the transcribed text, resulting in refined transcripts: $T_{asr} = Summarizer(Whisper(U))$.

\paragraph{Interleaved Chain-of-Thought.}
Vanilla CoT paradigm typically performs reasoning in one pass using only the input and a fixed instruction (i.e., `Let’s think step by step'). In contrast, our ViS-CoT framework adopts an interleaved reasoning procedure, where the model first performs a visual-only reasoning pass and then refines this reasoning by injecting auxiliary textual knowledge.
We begin by generating an initial reasoning path based solely on the video key frames $F_{p}^{key}$, the attribute definition $A_{def}$, and the base prompt $P_v$:
\begin{equation}
    \mathcal{R}^{'}_{ini}=\underset{\mathcal{R}^{'}}{argmax}P(\mathcal{R}^{'}|F_{p}^{key}, A_{def}, P_{v})
\end{equation}
This initial reasoning $\mathcal{R}'_{ini}$ reflects what the model can infer solely from visual evidence, without external context. Next, we refine this visual-only reasoning by incorporating captions $D_p$, refined speech transcripts $T_{asr}$, and retrieved product knowledge $\{P_i\}_{i=1}^{k}$. Instead of concatenating everything at once, the model revisits and updates its earlier reasoning in light of the new textual cues:

\begin{equation}
\begin{aligned}
\mathcal{R}_{p}
= \arg\max_{\mathcal{R}}\;
P\big(
&\mathcal{R} \mid \mathcal{R}'_{ini}, A_{def}, P_v, \\
&F_{p}^{key} \oplus D_p \oplus T_{asr}
\oplus \{P_i\}_{i=1}^{k}
\big).
\end{aligned}
\end{equation}


where $\mathcal{R}_{p}$ represents the refined reasoning path, in which auxiliary information is incrementally incorporated to resolve ambiguity, enrich attribute interpretation, and ground the reasoning more reliably across modalities.

\subsection{Integration and Generation}
After collecting various forms of auxiliary textual information, we align and integrate them into a unified input. Specifically, we combine captions $D_{p}$, refined transcripts $T_{asr}$, similar product data $\left\{P_{1}, \cdots, P_{k} \right\}$, and the interleaved reasoning outcome $R_{p}$. These enriched textual inputs are then fused in chronological order, combined with the visual evidence from video key frames $F_{p}^{key}$, to jointly encode semantic and relational knowledge.
This input is then fed into the video vision language models to generate the final result. The overall process can be formulated as:
\begin{equation}
    y = VLM(F_{p}^{key}, D_{p} \oplus \left\{P_{1}, \cdots, P_{k} \right\} \oplus T_{asr} \oplus R_{p})
\end{equation}
where $y$ denotes the final output. This integration preserves complementary multimodal cues within a unified representation, enabling the video VLM to ground its reasoning in richer and more diverse evidence while simultaneously reducing ambiguity.

\section{Experiments}
\subsection{Experimental Setup}
\paragraph{Dataset and Baselines.}
We evaluate ViS-CoT on a public dataset \textit{VideoAVE}~\cite{10.1145/3746252.3761621}, which is a video-to-text multi-attribute AVE dataset across 14 different categories with 172 unique attributes in e-Commerce. 
We evaluate our ViS-CoT in five open-source Video-Language Models, including Video-LLaVA (Video-LLaVA-7B)~\cite{lin-etal-2024-video}, VideoLLaMA3 (VideoLLaMA3-7B)~\cite{damonlpsg2025videollama3}, InternVideo2.5 (InternVideo2\_5\_Chat\_8B)~\cite{wang2025internvideo}, InternVL3.5 (InternVL3\_5-8B-Instruct)~\cite{wang2025internvl3_5}, and Qwen2.5-VL (Qwen2.5-VL-7B)~\cite{Qwen2.5-VL}.


\begin{table*}[]
\tiny
\setlength{\tabcolsep}{5pt}
\centering
\caption{\label{tab:main_result} Category-level fuzzy F1 (\%) under the attribute-conditioned and generalized settings. Best results are highlighted in bold, and models with a light blue background denote the base model augmented with the ViS-CoT.}
\begin{tabular}{lccccccccccccccc}
\hline
Model                    & Appl. & Arts  & Auto  & Baby  & Beauty & Clothes & Groc. & Indus. & Music & Patio & Pet   & Phones & Sports & Toys  & Gains \\ \hline
\multicolumn{16}{c}{Attribute-Conditioned Setting}                                                                                                     \\ \hline
VideoLLaVA               & 15.96 & 11.68 & 13.26 & 12.53 & 10.97  & 19.73   & 10.16 & 14.04  & 15.16 & 12.93 & 15.09 & 13.02  & 14.56  & 15.26 & -     \\
\rowcolor{lightblue}
VideoLLaVA + ViS-CoT     & 29.27 & 23.58 & 21.30 & 33.97 & 25.54  & 42.31   & 31.84 & 29.56  & 32.42 & 28.97 & 23.17 & 25.16  & 29.64  & 29.93 & \textcolor{red}{\,↑15.17}     \\
VideoLLaMA3              & 33.33 & 28.72 & 29.22 & 33.44 & 29.20  & 35.93   & 25.97 & 31.35  & 31.84 & 29.56 & 29.49 & 30.34  & 31.89  & 33.10 & -     \\
\rowcolor{lightblue}
VideoLLaMA3 + ViS-CoT    & 41.80 & 33.62 & 36.22 & 48.19 & 40.98  & 55.11   & 44.25 & 37.19  & 42.16 & 42.66 & 41.92 & 35.63  & 36.06  & 37.94 & \textcolor{red}{\,↑10.03} \\
InternVideo2.5           & 39.22 & 31.67 & 32.22 & 35.22 & 32.01  & 34.93   & 26.87 & 35.24  & 37.46 & 32.18 & 30.82 & 35.17  & 33.61  & 33.94 & -     \\
\rowcolor{lightblue}
InternVideo2.5 + ViS-CoT & 57.49 & 43.81 & 49.43 & 58.12 & 47.00  & 55.16   & 38.37 & 54.23  & 58.21 & 56.41 & 46.77 & 46.58  & 57.91  & 52.73 &    \textcolor{red}{\,↑17.98}   \\
InternVL3.5              & 26.79 & 27.41 & 28.18 & 33.91 & 29.02  & 26.88   & 23.03 & 30.37  & 36.66 & 28.11 & 28.96 & 32.89  & 31.69  & 32.14 & -     \\
\rowcolor{lightblue}
InternVL3.5 + ViS-CoT    & 58.49 & \textbf{51.12} & \textbf{52.40} & 62.50 & \textbf{55.56}  & 62.73   & 55.40 & \textbf{57.79}  & \textbf{63.75} & 58.60 & \textbf{53.80} & \textbf{54.11}  & 63.15  & \textbf{59.06} &   \textcolor{red}{\,\textbf{↑28.03}}    \\
Qwen2.5-VL               & 39.18 & 31.58 & 32.40 & 37.63 & 33.23  & 42.51   & 31.01 & 34.27  & 37.75 & 33.08 & 30.88 & 35.19  & 36.40  & 36.86 & -     \\
\rowcolor{lightblue}
Qwen2.5-VL + ViS-CoT     & \textbf{63.24} & 49.07 & 52.10 & \textbf{63.64} & 52.88  & \textbf{67.90}   & \textbf{56.54} & 51.85  & 62.12 & \textbf{61.32} & 52.99 & 50.26  & \textbf{66.87}  & 57.89 &    \textcolor{red}{\,↑22.62}      \\ \hline
\multicolumn{16}{c}{Generalized Setting}                                                                                                               \\ \hline
VideoLLaVA               & 5.65  & 5.36  & 6.47  & 6.95  & 4.69   & 9.13    & 3.27  & 6.48   & 6.09  & 6.27  & 4.70  & 5.81   & 7.21   & 5.76  & -     \\
\rowcolor{lightblue}
VideoLLaVA + ViS-CoT     & 20.02 & 18.79 & 17.88 & 21.06 & 14.52  & 32.70   & 11.52 & 23.62  & 25.27 & 22.42 & 15.36 & 22.79  & 25.57  & 18.86 &    \textcolor{red}{\,↑14.75}       \\
VideoLLaMA3              & 9.23  & 8.70  & 8.14  & 11.04 & 8.03   & 12.35   & 4.75  & 8.96   & 8.33  & 9.18  & 7.84  & 10.06  & 10.59  & 8.30  & -     \\
\rowcolor{lightblue}
VideoLLaMA3 + ViS-CoT    & 27.36 & 24.18 & \textbf{23.95} & \textbf{32.45} & \textbf{27.09}  & \textbf{37.58}   & \textbf{27.41} & 29.13  & 29.75 & \textbf{28.25} & \textbf{24.35} & \textbf{26.15}  & 28.80  & 24.55 &    \textcolor{red}{\textbf{\,↑18.96}}          \\
InternVideo2.5           & 11.95 & 12.32 & 10.16 & 14.76 & 11.37  & 16.40   & 8.22  & 11.89  & 9.93  & 10.73 & 9.07  & 14.51  & 13.29  & 9.73  & -     \\
\rowcolor{lightblue}
InternVideo2.5 + ViS-CoT & \textbf{28.59} & \textbf{24.52} & 23.49 & 30.71 & 25.72  & 36.55   & 23.66 & \textbf{29.37}  & \textbf{32.64} & 27.90 & 23.74 & 25.85  & 28.68  & \textbf{25.85} &  \textcolor{red}{\,↑15.92}       \\
InternVL3.5              & 7.05  & 8.43  & 6.25  & 9.58  & 6.49   & 11.35   & 3.59  & 7.71   & 8.02  & 7.05  & 5.20  & 9.70   & 8.54   & 6.94  &  -     \\
\rowcolor{lightblue}
InternVL3.5 + ViS-CoT    & 23.99 & 20.66 & 20.35 & 27.51 & 22.30  & 30.16   & 21.11 & 24.99  & 30.02 & 24.60 & 22.14 & 22.28  & 28.36  & 21.70 &     \textcolor{red}{\,↑16.73}        \\
Qwen2.5-VL               & 4.66  & 7.24  & 5.61  & 9.27  & 6.47   & 10.79   & 6.15  & 5.64   & 8.33  & 6.21  & 5.05  & 10.06  & 7.82   & 5.65  & -     \\
\rowcolor{lightblue}
Qwen2.5-VL + ViS-CoT     & 26.68 & 21.50 & 20.76 & 30.06 & 25.29  & 33.23   & 24.97 & 24.66  & 30.55 & 24.76 & 22.66 & 25.32  & \textbf{29.27}  & 24.22 &   \textcolor{red}{\,↑18.93}           \\\hline
\end{tabular}
\end{table*}

\paragraph{Evaluation Metrics.}
We follow the evaluation criteria of~\cite{10.1145/3746252.3761621} to evaluate VideoAVE from two perspectives: (1) attribute-conditioned value prediction, where a predefined list of attributes is provided, and (2) open attribute-value pair extraction, where no attributes are given in advance. 
Consistent with prior works~\cite{zhang-etal-2023-pay, gong-etal-2025-visual, 10.1145/3746252.3761621}, we adopt the fuzzy F1 score as our evaluation metric. 
We report results at two levels: category-level, where fuzzy F1 is aggregated across all attribute-value predictions within each product category, and attribute-level, where fuzzy F1 is computed separately for each attribute.

\paragraph{Implementation Details.}
\label{sec:appendix_implementation}
We performed all experiments on NVIDIA A100 GPUs with 80GB memory.
All the base models take the default configurations without additional modification. For ViS-CoT, frame quality filtering is applied with thresholds set to $\tau_{\mathrm{blur}}=100$, $\tau_{\mathrm{black}}=10$, $\tau_{\mathrm{white}}=245$, and $\tau_{\mathrm{contrast}}=15$. In visual clustering, the number of clusters is fixed to $m=5$, while the top-$k$ retrieved items in the visual search stage is set to $k=10$. The influence of both $m$ and $k$ is analyzed in Sec.~\ref{sec:parameter}.
For the interleaved CoT, the captioning module ($Caption$) uses Qwen2.5-VL while the summarization module ($Summarizer$) and the attribute definition generator ($A_{def}$) are based on GPT-4.1.
The final prompt provided to any video VLM in the attribute-conditioned setting follows the template: 
\textit{...Extract the values for these attributes: \colorbox{softgray}{attributes} in the images given the following additional context: \colorbox{softpink}{Captions $D_{p}$}, \colorbox{softgreen}{Knowledge ${P_{1}, \cdots, P_{k}}$}, \colorbox{softblue}{Transcripts $T_{asr}$}, \colorbox{softorange}{Reasoning $R_{p}$}. Choose one specific value for each attribute based on your analysis...}.
For the general setting, the instruction “\textit{Extract the values...}” is replaced with “\textit{Given the following rules: \colorbox{softgray}{rules}, generate the most relevant attributes based on what you can observe.}”

Additional analyses are provided in the \textbf{Appendix}, including 
latency evaluation (Appendix~\ref{sec:appendix_latency}), ablations (Appendix~\ref{sec:appendix_ablation}), single-pass vs. interleaved comparison (Appendix~\ref{sec:appendix_singlepass}), label-leakage robustness (Appendix~\ref{sec:appendix_leakage}), and open-source reproducibility (Appendix~\ref{sec:appendix_opensource}).

\begin{table*}[]
\tiny
\setlength{\tabcolsep}{3.4pt}
\centering
\caption{\label{tab:attribute_result} Attribute-level fuzzy F1 (\%) results. Top 10 frequent attributes are reported. Best results are highlighted in bold, and models with a light green background denote the base model augmented with the plug-and-play ViS-CoT.}
\begin{tabular}{lcccccccccc}
\hline
Model                & Brand & Color & Material & \begin{tabular}[c]{@{}c@{}}Country\\ of Origin\end{tabular} & \begin{tabular}[c]{@{}c@{}}Item\\ Form\end{tabular} & \begin{tabular}[c]{@{}c@{}}Power\\ Source\end{tabular} & Shape & \begin{tabular}[c]{@{}c@{}}Suggested\\ Users\end{tabular} & Pattern & \begin{tabular}[c]{@{}c@{}}Finish\\ Type\end{tabular} \\ \hline
VideoLLaVA           & 6.49  & 19.62 & 15.12    & 15.61                                                       & 13.02                                               & 2.31                                                   & 22.89 & 2.28                                                      & 21.85   & 10.68                                                 \\
\rowcolor{lightgreen}
VideoLLaVA + ViS-CoT & 28.67\textsubscript{\tiny\textcolor{red}{↑22.18}} & 22.44\textsubscript{\tiny\textcolor{red}{↑2.82}}
& 31.02 \textsubscript{\tiny\textcolor{red}{↑15.90}}   & 21.34 \textsubscript{\tiny\textcolor{red}{↑5.73}}                                                      & 33.19  \textsubscript{\tiny\textcolor{red}{↑20.17}}                                          & 41.64           \textsubscript{\tiny\textcolor{red}{↑39.33}}                                     & 31.31\textsubscript{\tiny\textcolor{red}{↑8.42}} & 42.55 \textsubscript{\tiny\textcolor{red}{↑40.27}}                                                  & 29.38 \textsubscript{\tiny\textcolor{red}{↑7.53}} & 16.65    \textsubscript{\tiny\textcolor{red}{↑5.97}}                                              \\VideoLLaMA3           & 43.27 & 32.76 & 28.34    & 29.68                                                       & 19.82                                               & 8.82                                                   & 34.70  & 1.68                                                      & 27.64   & 14.52                                                 \\
\rowcolor{lightgreen}
VideoLLaMA3 + ViS-CoT & 34.90\textsubscript{\tiny\textcolor{ForestGreen}{↓8.37}}    & 28.84\textsubscript{\tiny\textcolor{ForestGreen}{↓3.92}}   & 37.08\textsubscript{\tiny\textcolor{red}{↑8.74}}      & 47.06\textsubscript{\tiny\textcolor{red}{↑17.38}}                                                         & 44.03\textsubscript{\tiny\textcolor{red}{↑24.21}}                                                 & 58.83 \textsubscript{\tiny\textcolor{red}{↑50.01}}                                                   & 56.45\textsubscript{\tiny\textcolor{red}{↑21.75}}   & 45.57\textsubscript{\tiny\textcolor{red}{↑43.89}}                                                       & 56.25\textsubscript{\tiny\textcolor{red}{↑28.61}}     & 32.42\textsubscript{\tiny\textcolor{red}{↑17.90}}                                                   \\InternVideo2.5           & 42.41 & 34.95 & 28.37    & 43.78                                                       & 19.45                                               & 8.03                                                   & 29.95 & 1.58                                                      & 44.13   & 17.47                                                 \\
\rowcolor{lightgreen}
InternVideo2.5 + ViS-CoT & 42.12\textsubscript{\tiny\textcolor{ForestGreen}{↓0.29}}   & 39.71\textsubscript{\tiny\textcolor{red}{↑4.76}}   & 51.33\textsubscript{\tiny\textcolor{red}{↑22.96}}      & \textbf{75.21}\textsubscript{\tiny\textcolor{red}{↑31.43}}                                                         & 47.90\textsubscript{\tiny\textcolor{red}{↑28.45}}                                                  & 70.69\textsubscript{\tiny\textcolor{red}{↑62.66}}                                                    & 63.08\textsubscript{\tiny\textcolor{red}{↑33.13}}   & 80.21\textsubscript{\tiny\textcolor{red}{↑78.63}}                                                       & 68.94\textsubscript{\tiny\textcolor{red}{↑24.81}}     & 40.02\textsubscript{\tiny\textcolor{red}{↑22.55}}                                                   \\
            InternVL3.5              & 46.91 & 33.13 & 27.77    & 36.87                                                       & 14.63                                               & 13.87                                                  & 37.87 & 2.10                                                      & 48.25   & 14.43                                                 \\
\rowcolor{lightgreen}
InternVL3.5 + ViS-CoT    & \textbf{50.40}\textsubscript{\tiny\textcolor{red}{↑3.49}} & \textbf{43.43}\textsubscript{\tiny\textcolor{red}{↑10.30}} & 56.45\textsubscript{\tiny\textcolor{red}{↑28.68}}    & 74.60\textsubscript{\tiny\textcolor{red}{↑37.73}}                                                       & \textbf{57.24}\textsubscript{\tiny\textcolor{red}{↑42.61}}                                               & \textbf{74.29}\textsubscript{\tiny\textcolor{red}{↑60.42}}                                                  & \textbf{69.46}\textsubscript{\tiny\textcolor{red}{↑31.59}} & \textbf{86.64}\textsubscript{\tiny\textcolor{red}{↑84.54}}                                                     & \textbf{74.40}\textsubscript{\tiny\textcolor{red}{↑26.15}}   & \textbf{47.27}\textsubscript{\tiny\textcolor{red}{↑32.84}}                                                 \\Qwen2.5-VL               & 46.89 & 34.42 & 33.69    & 34.90                                                       & 25.34                                               & 11.96                                                  & 35.71 & 1.87                                                      & 46.16   & 16.27                                                 \\
\rowcolor{lightgreen}
Qwen2.5-VL + ViS-CoT     & 48.34\textsubscript{\tiny\textcolor{red}{↑1.45}}  & 39.54\textsubscript{\tiny\textcolor{red}{↑5.12}}  & \textbf{59.37}\textsubscript{\tiny\textcolor{red}{↑25.68}}     & 73.15\textsubscript{\tiny\textcolor{red}{↑38.25}}                                                        & 53.50\textsubscript{\tiny\textcolor{red}{↑28.16}}                                                & 73.55\textsubscript{\tiny\textcolor{red}{↑61.59}}                                                   & 68.77\textsubscript{\tiny\textcolor{red}{↑33.06}}  & 79.87\textsubscript{\tiny\textcolor{red}{↑78.00}}                                                      & 66.56\textsubscript{\tiny\textcolor{red}{↑20.40}}    & 42.14\textsubscript{\tiny\textcolor{red}{↑25.87}}                                                  \\ \hline
\end{tabular}
\end{table*}

\subsection{Results and Discussions}
\subsubsection{Main Results}

\textit{Category-Level Results.}
Table \ref{tab:main_result} reports category-level performance. We observe that (1) Integrating ViS-CoT consistently improves all base VLMs across both attribute-conditioned and generalized settings, highlighting ViS-CoT’s effectiveness and generality as a plug-and-play module that boosts SOTA video-to-text AVE without fine-tuning.
(2) ViS-CoT is particularly effective in categories that are visually complex and semantically rich, suggesting that the integration of visual clustering, retrieved product context, and CoT reasoning is especially valuable when raw videos alone are insufficient for precise extraction.
(3) Performance under the attribute-conditioned setting is generally higher where the attributes are provided. We notice that even in the more challenging generalized setting, ViS-CoT also demonstrates consistent improvements, highlighting the robustness of ViS-CoT in open-world scenarios where attributes should be discovered automatically.

\textit{Attribute-Level Results.} Table~\ref{tab:attribute_result} reports attribute-level performance. Due to space limitations, we present the 10 most frequent attribute results. We observe: (1) ViS-CoT consistently improves performance across fine-grained visual and semantic attributes. 
For example, `power source' may only be inferred from some specific usage scenes, while `item form' may be revealed during product interactions rather than from static packaging. Such attributes require reasoning over complementary evidence, which ViS-CoT enables through interleaved visual-textual cues.
(2) Some attributes, such as `country of origin' and `suggested users' are often ambiguous or absent from visual appearance. In these cases, retrieved similar product information significantly improves the performance, confirming that the retrieved knowledge is essential for filling gaps not inferable from video alone.
(3) For some relatively `easy' attributes such as `brand' and `color', ViS-CoT may cause a negative effect.
We think that these attributes are readily identifiable from almost any frame.
The added complexity of reasoning can introduce noise or hallucinations. Simplified strategies can be explored in future work to balance reasoning depth with attribute-specific difficulty.

\subsubsection{Ablation Study}\label{sec:ablation}
\begin{figure}[t!] 
 \center{\includegraphics[height=6cm,width=7.5cm]{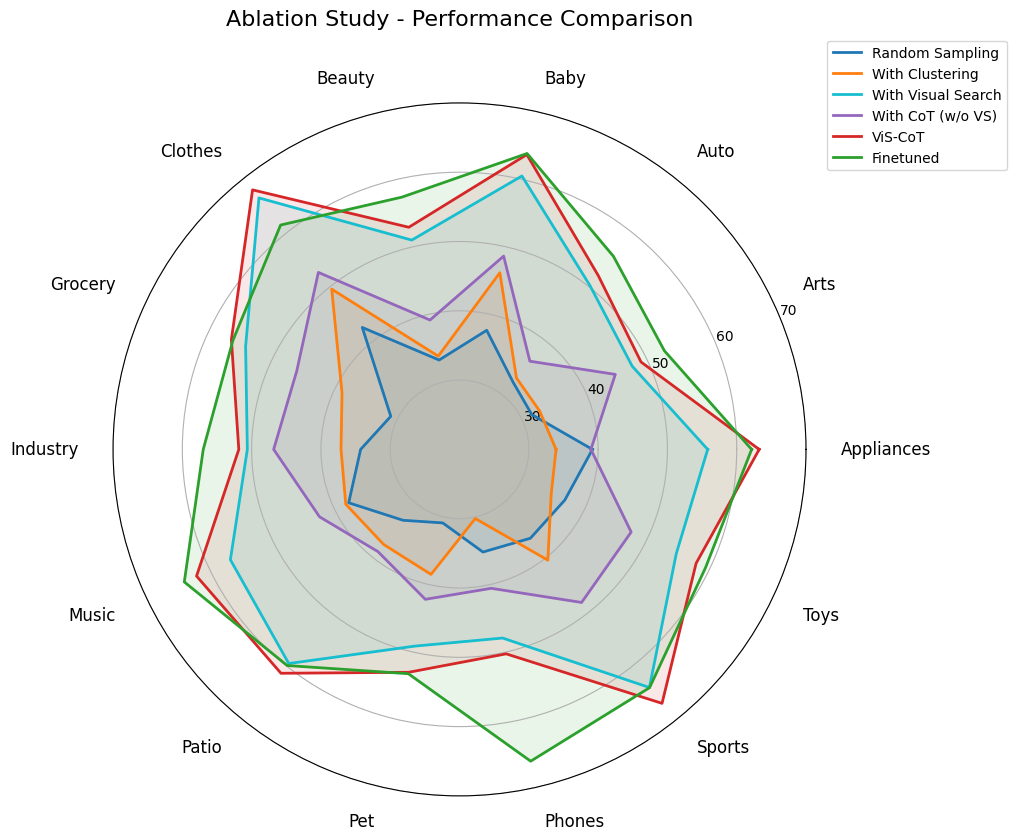}}
 \caption{\label{fig:ablation} Performance comparison (micro-F1\%) of frames with random sampling, key frame clustering, visual search augmentation, CoT reasoning, ViS-CoT with search-enhanced interleaved CoT, and a full-size finetuned base model using training data.}
 \vspace{-4mm}
 \end{figure}

 \begin{figure*}[t]
  \centering
  \includegraphics[width=\textwidth,keepaspectratio]{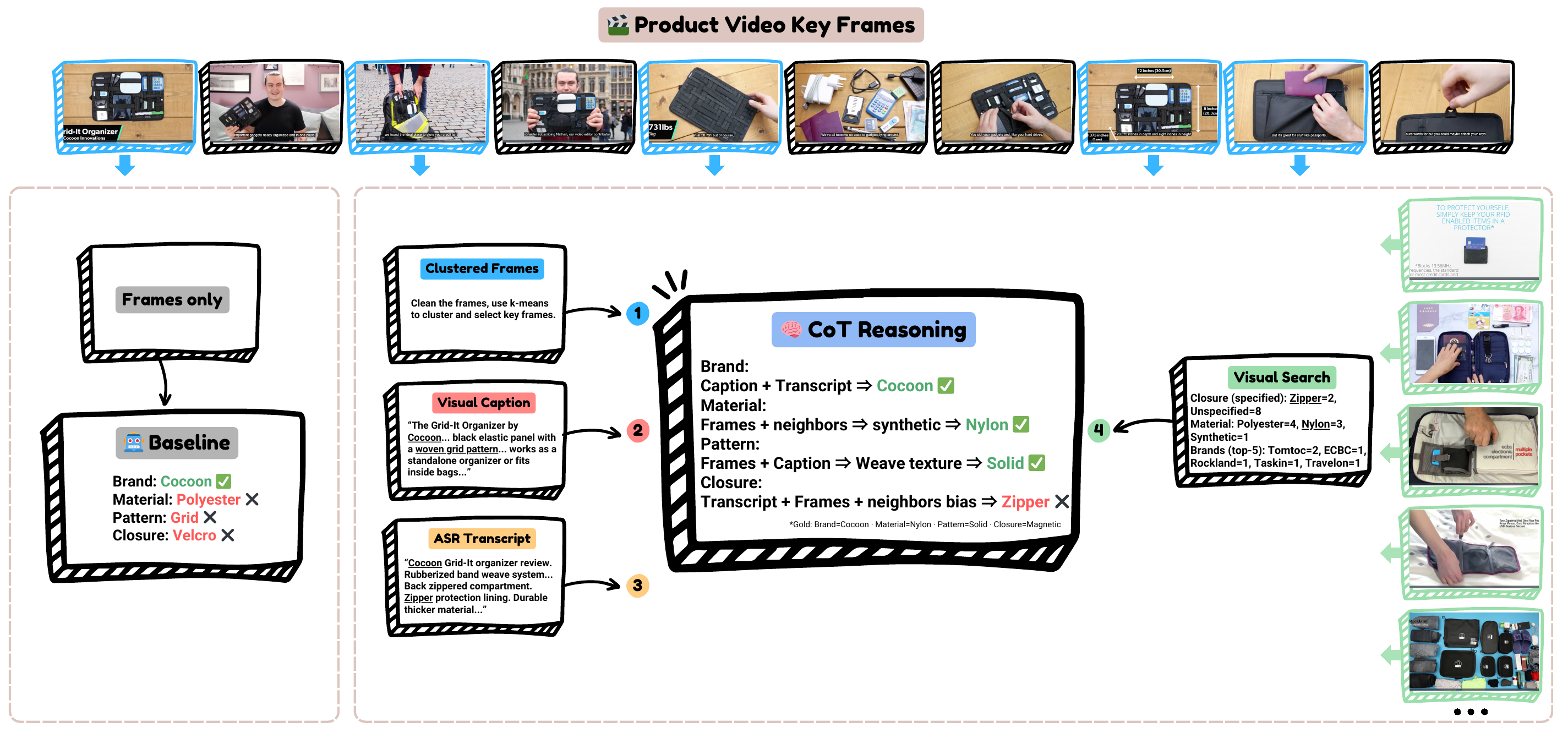}%
  \caption{\label{fig:case}
\textbf{Case study on a Cocoon organizer sleeve.} 
Top: product video key frames. \textcolor{black}{\textbf{black}} border: original extracted frames; \textcolor{blue}{\textbf{blue}} border: clustered/selected key frames.
\textbf{Left panel (Baseline):} uniformly sampled video frames, without key-frame selection. 
\textbf{Right panel (ViS-CoT):} 
(1) clustered \& cleaned key frames,
(2) visual caption from the selected frames,
(3) ASR transcript (speaker mentions brand, functions, and details),
(4) visual-search neighbors providing attribute priors; 
Fusing (1)–(4) corrects Material$\rightarrow$Nylon and Pattern$\rightarrow$Solid; Closure remains biased to Zipper (gold: Magnetic).}
  \vspace{-2mm}
\end{figure*}

Our ablation study isolates the effect of each key module and quantifies its impact on overall end-to-end performance.
We also compare the performance against a full-size finetuned VLM trained on the whole training data.
All experiments are conducted using the same Qwen2.5-VL-7B-Instruct base model for fair comparison. 
From Figure~\ref{fig:ablation}, we observe: 
(1) Clustering proves more effective than random sampling for frame selection. Selling videos often include background elements or non-relevant items, and visual clustering emphasizes the most representative product-centric frames, reducing redundant and noisy information.
(2) Visual search augmentation in ViS-CoT outperforms purely text-based reasoning. By retrieving visually similar products, the model gains access to items that often share key attributes with the query product. This retrieved information can further correct errors that may arise in subsequent interleaved CoT step. 
(3) By leveraging visual cues to guide and refine the CoT, ViS-CoT achieves the best performance across almost all categories. This training-free framework achieves competitive performance compared with the full-size finetuned VLM, highlighting the promise of retrieval-augmented reasoning as a scalable and efficient paradigm for video-to-text AVE in e-Commerce.
(4) A single-pass variant concatenating all auxiliary signals reaches only 46.5\% F1, 11.9 pp below full ViS-CoT and even below visual search alone, confirming the staged design is essential (Appendix~\ref{sec:appendix_singlepass}). 
(5) On a clean test subset removing all training-overlapping products, visual search retains a +13.3 pp gain (vs. +14.9 pp on the full set), confirming gains stem from genuine visual similarity, not label leakage (Appendix~\ref {sec:appendix_leakage}).

\subsubsection{Case Study}
To illustrate how ViS-CoT improves AVE, we analyze a Cocoon organizer sleeve (Figure~\ref{fig:case}). The \emph{baseline} consumes only video frames and gets only `Brand: Cocoon' correct, while misclassifying Material as Polyester, Pattern as Grid, and Closure as Velcro. In contrast, \emph{ViS-CoT} augments the clustered key frames with caption, ASR transcript, and top-$k$ visual-search products, then reasons over these signals. Close-ups of the weave with retrieved similar information corrects it to \emph{Nylon}; the `textured' cue in captions disambiguates weave texture from a printed pattern, yielding \emph{Solid}. Transcript phrases (`back \emph{zippered} compartment,' `\emph{zipper} protection lining') still pull the chain toward \emph{Zipper}, whereas the video's main flap is magnetic, an informative, auditable residual error. 

\subsubsection{Parameter Sensitivity}\label{sec:parameter}
\begin{figure}[t!] 
 \center{\includegraphics[height=4cm,width=7.5cm]{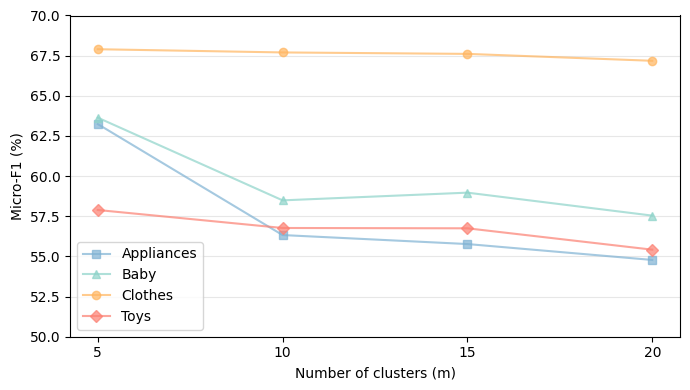}}
 \caption{\label{fig:cluster} Performance with different cluster number $m$ in visual clustering on four categories of VideoAVE.}
 \vspace{-4mm}
 \end{figure}

\paragraph{Effect of different cluster number $m$.}
To analyze the effect of the number of clusters $m$, we select four representative categories that vary in dataset size, domain characteristics, and clustering sensitivity. We evaluate $m \in \left\{5,10,15,20 \right\}$ in the attribute-conditioned setting.
As shown in Figure~\ref{fig:cluster}, increasing $m$ generally leads to a slight decline in performance. 
This suggests that using fewer clusters can yield better results, as selecting a moderate number (i.e. $m=5$) helps the model focus on semantically coherent groups of attributes, balancing granularity and stability. 
While finer clustering can capture more detailed information, excessive cluster numbers may introduce irrelevant product information, causing over-fragmentation and reducing overall performance.

\begin{figure*}[t!] 
 \center{\includegraphics[height=6.5cm,width=\textwidth]{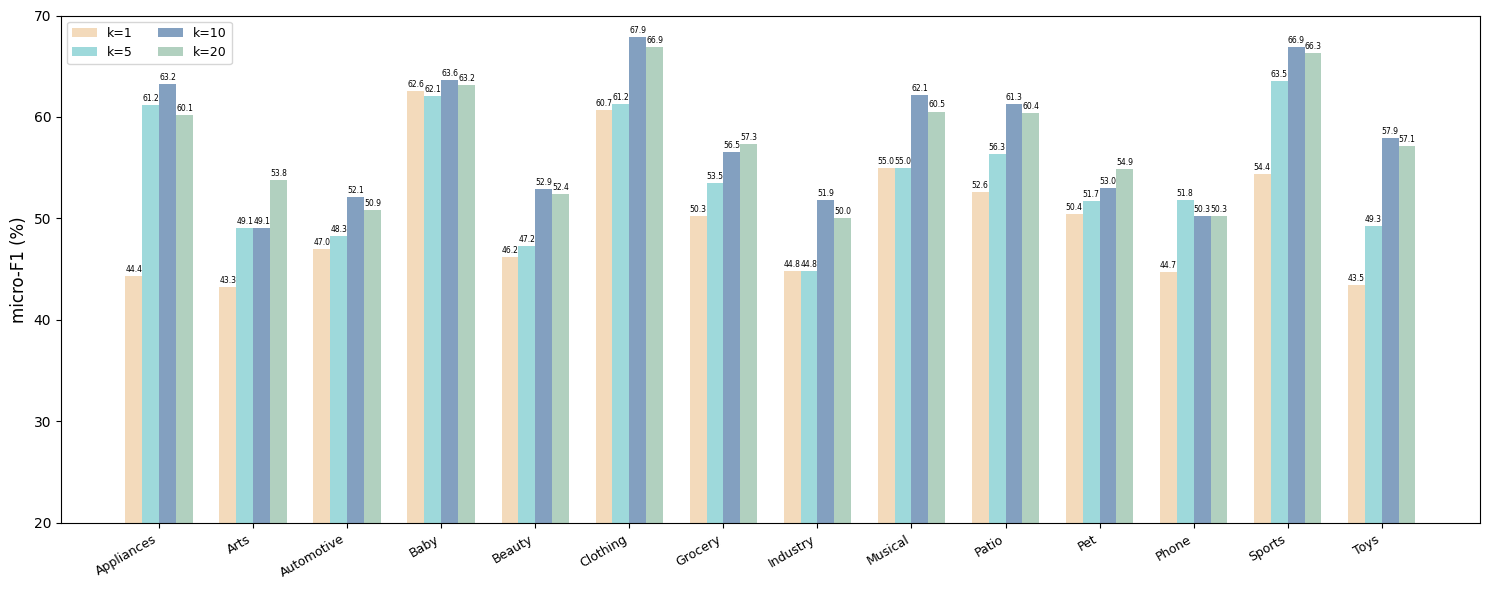}}
 \caption{\label{fig:cluster_bars} Performance with different number of top-$k$ in visual search when using Qwen2.5-VL-7B as the base.}
 \vspace{-4mm}
 \end{figure*}

\paragraph{Effect of different number of top-$k$.}

To explore top-$k$ retrieval effect in the visual search component, we evaluate $k \in \left\{1,5,10,20 \right\}$  in the attribute-conditioned setting, with results shown in Figure~\ref{fig:cluster_bars}. As illustrated, $k=10$ provides the best overall performance, achieving the highest micro-F1 in most categories.
Though $k=20$ shows slightly better performance in categories of Arts, Grocery, and Pet, it often introduces redundant or noisy information that negatively impacts other categories and increases latency during inference.
Thus, we adopt $k=10$ in our final configuration, which best balances effectiveness and efficiency.

\subsubsection{Error Analysis}
\label{sec:appendix_error}

To better understand the limitations of ViS-CoT, we analyze prediction errors produced by the full system across seven representative categories.

\paragraph{Failure Mode Taxonomy.}
Based on manual inspection of incorrect predictions, we group errors into three categories:

\begin{itemize}
    \item \textbf{Type I: Visually Ambiguous} (28.6\%) — Attributes that are difficult to determine from the available video frames (e.g., \textit{material}, \textit{country of origin}, \textit{item form}). In these cases, the visual evidence is insufficient to uniquely identify the correct value.
    
    \item \textbf{Type II: Knowledge Dependent} (28.7\%) — Attributes that require external or domain-specific knowledge not present in the video or among retrieved neighbors (e.g., niche \textit{brand} identities or \textit{compatible devices}).
    
    \item \textbf{Type III: Hallucination} (42.7\%) — The model generates a plausible but incorrect value that is not supported by the available visual or textual evidence. These errors typically occur when multiple candidate values are possible.
\end{itemize}

The distribution of failure modes varies across categories. 
Hallucination errors are the most frequent type overall, accounting for approximately 37--50\% of failures across the seven evaluated categories. 
Knowledge-dependent errors occur more frequently in Pet (57\%) and Toys (48\%), which contain attributes requiring specialized product knowledge (e.g., \textit{compatible device type}, \textit{assembly required}). 
Visually ambiguous errors appear more often in Patio (30\%) and Grocery (29\%), where attributes such as \textit{material grade} or \textit{item form} are difficult to determine from visual cues alone.

These observations suggest several possible directions for improving performance, including incorporating additional structured product knowledge sources, adapting reasoning depth to attribute difficulty, and improving confidence calibration to reduce unsupported predictions.

\section{Conclusion}
We present ViS-CoT for video-to-text e-Commerce product AVE.
ViS-CoT leverages visual clustering to extract key frames from the video and integrates the retrieved similar products data, captions, transcripts, and interleaved CoT reasoning to enhance video understanding.
Unlike fine-tuning VLMs, which is resource-intensive and time-consuming, ViS-CoT provides a resource-efficient, plug-and-play solution that can be seamlessly applied to any open-source video VLMs.
ViS-CoT shows strong performance gains across five SOTA video VLMs under both attribute-conditioned and open AVE settings.

\section{Limitations}
While ViS-CoT demonstrates substantial improvements, several limitations remain.
First, although the multi-stage ViS-CoT pipeline yields an average micro-F1 gain of 17.91 percentage points, its iterative search and reasoning procedure introduces additional inference latency. This may hinder scalability in real-time or large-scale production environments. The inference latency is described and analyzed in the Appendix (Appendix~\ref{sec:appendix_latency}).
Second, ViS-CoT is specifically designed for product-focused video attribute value extraction, where structured aspect accuracy is prioritized. Consequently, we do not include comparisons with chain-of-thought approaches developed for broader video understanding tasks (e.g., storytelling, captioning, summarization). Such evaluations would require qualitatively different objectives and metrics and thus fall outside the scope of this work.
Third, our empirical evaluation is focused on video-to-text e-Commerce AVE, and VideoAVE~\cite{10.1145/3746252.3761621}, to the best of our knowledge, is the only publicly available benchmark directly aligned with this task. While this scope matches the problem targeted by ViS-CoT, future work should evaluate the framework on additional product-video datasets once comparable public resources become available. We mainly evaluate open-source video VLMs to support reproducibility, controllability, and cost-efficient large-scale experimentation. Although commercial multimodal systems such as GPT and Gemini are important future baselines, systematic evaluation with them is constrained by API cost, limited transparency, and changing model versions. Moreover, our Appendix~\ref{sec:appendix_opensource} shows that replacing GPT-4.1-based helper modules with open-source Qwen2.5-7B yields comparable performance, further supporting the practicality of the open-source setting.


\bibliography{custom}

\appendix


\section{Inference Latency Analysis}
\label{sec:appendix_latency}
\FloatBarrier

\begin{table}[h]
\centering
\small
\tabcolsep=0.13cm
\begin{tabular}{lcc}
\hline
\textbf{Component} & \textbf{Latency (s)} & \textbf{\%} \\
\hline
Keyframe clustering (subtotal) & 2.40 & 48.1 \\
\rowcolor{softgray}
\quad SigLIP embedding & 2.24 & (44.9) \\
\rowcolor{softgray}
\quad K-means & 0.16 & (3.2) \\
Visual search (FAISS) & 0.11 & 2.2 \\
CoT Stage 1 (vision hypothesis) & 0.97 & 19.4 \\
CoT Stage 2 (grounded extraction) & 1.51 & 30.3 \\
\hline
\rowcolor{lightblue}
\textbf{Full ViS-CoT total} & \textbf{4.99} & \textbf{100.0} \\
\hline
\textit{Baseline VLM (reference)} & \textit{1.31} & --- \\
\hline
\end{tabular}
\caption{Inference latency breakdown of ViS-CoT on a representative toy product. Indented gray rows break down the keyframe-clustering subtotal. Keyframe clustering is a one-time preprocessing step per video. Excluding it, per-query inference requires only 2.59~s vs.\ 1.31~s for the baseline.}
\label{tab:latency_breakdown}
\end{table}

\begin{table}[h]
\centering
\small
\tabcolsep=0.13cm
\begin{tabular}{lcccc}
\hline
\textbf{Condition} & \textbf{F1 (\%)} & \textbf{$\Delta$F1} & \textbf{s/product} & \textbf{F1 eff.} \\
\hline
C0~~Baseline & 33.4 & --- & 1.6 & --- \\
C1~~+Visual Search & 48.3 & $+14.9$ & 2.8 & 12.3~pp/s \\
C2~~+CoT (w/o VS) & 33.5 & $+0.1$ & 3.2 & 0.1~pp/s \\
C3~~Full ViS-CoT & 58.5 & $+25.1$ & 3.3 & \textbf{14.3~pp/s} \\
\hline
\end{tabular}
\caption{Per-condition inference time vs.\ F1 gain. F1 efficiency = $\Delta$F1 per additional second over C0. Full ViS-CoT delivers the best return on compute: each added second yields 14.3~pp of F1 improvement.}
\label{tab:latency_comparison}
\end{table}

We measure per-sample inference latency on a single GPU using Qwen2.5-VL-7B. 
Video downloading and frame extraction are excluded from all measurements. 
As shown in Table~\ref{tab:latency_breakdown}, the full ViS-CoT pipeline processes one product in 4.99~s, with component latencies summing to 100\%.

Keyframe clustering (SigLIP embedding followed by K-means) accounts for 2.40~s (48.2\%) of the total runtime. 
However, this step is executed only once per product video regardless of the number of attribute queries, and therefore acts as a one-time preprocessing cost rather than a per-query inference overhead. 
Excluding this preprocessing stage, per-query inference requires 2.59~s, representing a +1.28~s increase over the 1.31~s baseline.

Visual search implemented with FAISS introduces negligible overhead (0.11~s, 2.2\%). 
Table~\ref{tab:latency_comparison} further reports the trade-off between latency and performance across ablation settings. 
The full ViS-CoT system achieves the highest efficiency, yielding 14.3 percentage points of F1 improvement per additional second of inference. 
In contrast, CoT alone (C2) provides only 0.1~pp/s, indicating that retrieval-based visual grounding substantially improves the effectiveness of the reasoning process.
At query time (i.e., after one-time keyframe clustering), end-to-end throughput on a single A100 GPU is approximately \textbf{0.30 products/sec/GPU}, acceptable for offline catalog enrichment and trivially scalable via data parallelism.

\section{Ablation Study: Component Contributions}
\label{sec:appendix_ablation}

Table~\ref{tab:component_ablation} presents a systematic evaluation of each component of ViS-CoT using Qwen2.5-VL-7B as the base model across seven representative product categories. 
We consider four incremental conditions: \textbf{C0} (baseline: images + attributes only, without auxiliary context or CoT reasoning), \textbf{C1} (C0 + Visual Search), \textbf{C2} (C0 + interleaved CoT only, without retrieval), and \textbf{C3} (Full ViS-CoT with all components). 
To ensure a controlled and fair comparison, all conditions use the same inference framework with identical prompting and image preprocessing. 
Absolute F1 values are anchored to the baseline and full-system performance reported in Table~\ref{tab:main_result}; the relative improvements within each table isolate individual component contributions.

\begin{table}[h]
\centering
\small
\tabcolsep=0.11cm
\begin{tabular}{lrrrr}
\hline
\textbf{Category} & \textbf{C0} & \textbf{C1 (+VS)} & \textbf{C2 (+CoT)} & \textbf{Full} \\
\hline
Arts     & 31.6 & 43.1$_{\textcolor{BrickRed}{\scriptsize\uparrow11.5}}$ & 31.7$_{\textcolor{BrickRed}{\scriptsize\uparrow0.1}}$ & 49.1$_{\textcolor{BrickRed}{\scriptsize\uparrow17.5}}$ \\
Baby     & 37.6 & 57.2$_{\textcolor{BrickRed}{\scriptsize\uparrow19.6}}$ & 39.6$_{\textcolor{BrickRed}{\scriptsize\uparrow2.0}}$ & 63.6$_{\textcolor{BrickRed}{\scriptsize\uparrow26.0}}$ \\
Clothing & 32.4 & 47.3$_{\textcolor{BrickRed}{\scriptsize\uparrow14.9}}$ & 33.3$_{\textcolor{BrickRed}{\scriptsize\uparrow0.9}}$ & 67.9$_{\textcolor{BrickRed}{\scriptsize\uparrow35.5}}$ \\
Grocery  & 31.0 & 47.9$_{\textcolor{BrickRed}{\scriptsize\uparrow16.9}}$ & 31.8$_{\textcolor{BrickRed}{\scriptsize\uparrow0.8}}$ & 56.5$_{\textcolor{BrickRed}{\scriptsize\uparrow25.5}}$ \\
Patio    & 33.1 & 50.0$_{\textcolor{BrickRed}{\scriptsize\uparrow16.9}}$ & 32.3$_{\textcolor{ForestGreen}{\scriptsize\downarrow0.8}}$ & 61.3$_{\textcolor{BrickRed}{\scriptsize\uparrow28.2}}$ \\
Pet      & 30.9 & 40.2$_{\textcolor{BrickRed}{\scriptsize\uparrow9.3}}$ & 28.6$_{\textcolor{ForestGreen}{\scriptsize\downarrow2.3}}$ & 53.0$_{\textcolor{BrickRed}{\scriptsize\uparrow22.1}}$ \\
Toys     & 36.9 & 52.2$_{\textcolor{BrickRed}{\scriptsize\uparrow15.3}}$ & 37.2$_{\textcolor{BrickRed}{\scriptsize\uparrow0.3}}$ & 57.9$_{\textcolor{BrickRed}{\scriptsize\uparrow21.0}}$ \\
\hline
\textbf{Avg} & \textbf{33.4} & \textbf{48.3}$_{\textcolor{BrickRed}{\scriptsize\uparrow14.9}}$ & \textbf{33.5}$_{\textcolor{BrickRed}{\scriptsize\uparrow0.1}}$ & \textbf{58.5}$_{\textcolor{BrickRed}{\scriptsize\uparrow25.1}}$ \\
\hline
\end{tabular}
\caption{Component ablation (micro-F1\%, $\uparrow$/$\downarrow$ vs.\ C0). C0: base VLM with images and attributes only; C1: C0 + Visual Search (RAG); C2: C0 + interleaved CoT only, no retrieval; Full (C3): complete ViS-CoT. Subscripts show absolute change from C0.}
\label{tab:component_ablation}
\end{table}

Table~\ref{tab:knockout_ablation} shows the contribution of each modality by removing it from the full system.

\begin{table}[h]
\centering
\small
\tabcolsep=0.10cm
\begin{tabular}{lrrrr}
\hline
\textbf{Category} & \textbf{Full} & \textbf{w/o Cap.} & \textbf{w/o Trans.} & \textbf{w/o RAG} \\
\hline
Arts     & 49.1 & 39.5$_{\textcolor{ForestGreen}{\scriptsize\downarrow9.6}}$ & 39.4$_{\textcolor{ForestGreen}{\scriptsize\downarrow9.7}}$ & 37.9$_{\textcolor{ForestGreen}{\scriptsize\downarrow11.2}}$ \\
Baby     & 63.6 & 57.5$_{\textcolor{ForestGreen}{\scriptsize\downarrow6.1}}$ & 58.0$_{\textcolor{ForestGreen}{\scriptsize\downarrow5.6}}$ & 54.5$_{\textcolor{ForestGreen}{\scriptsize\downarrow9.1}}$ \\
Clothing & 67.9 & 61.7$_{\textcolor{ForestGreen}{\scriptsize\downarrow6.2}}$ & 61.4$_{\textcolor{ForestGreen}{\scriptsize\downarrow6.5}}$ & 59.8$_{\textcolor{ForestGreen}{\scriptsize\downarrow8.1}}$ \\
Grocery  & 56.5 & 48.1$_{\textcolor{ForestGreen}{\scriptsize\downarrow8.4}}$ & 49.2$_{\textcolor{ForestGreen}{\scriptsize\downarrow7.3}}$ & 46.3$_{\textcolor{ForestGreen}{\scriptsize\downarrow10.2}}$ \\
Patio    & 61.3 & 54.4$_{\textcolor{ForestGreen}{\scriptsize\downarrow6.9}}$ & 54.4$_{\textcolor{ForestGreen}{\scriptsize\downarrow6.9}}$ & 51.8$_{\textcolor{ForestGreen}{\scriptsize\downarrow9.5}}$ \\
Pet      & 53.0 & 47.8$_{\textcolor{ForestGreen}{\scriptsize\downarrow5.2}}$ & 48.0$_{\textcolor{ForestGreen}{\scriptsize\downarrow5.0}}$ & 46.5$_{\textcolor{ForestGreen}{\scriptsize\downarrow6.5}}$ \\
Toys     & 57.9 & 51.0$_{\textcolor{ForestGreen}{\scriptsize\downarrow6.9}}$ & 50.6$_{\textcolor{ForestGreen}{\scriptsize\downarrow7.3}}$ & 48.4$_{\textcolor{ForestGreen}{\scriptsize\downarrow9.5}}$ \\
\hline
\textbf{Avg} & \textbf{58.5} & \textbf{51.4}$_{\textcolor{ForestGreen}{\scriptsize\downarrow7.1}}$ & \textbf{51.6}$_{\textcolor{ForestGreen}{\scriptsize\downarrow6.9}}$ & \textbf{49.3}$_{\textcolor{ForestGreen}{\scriptsize\downarrow9.2}}$ \\
\hline
\end{tabular}
\caption{Modality knock-out ablation (micro-F1\%, $\downarrow$ drop from Full). Cap.=visual captions; Trans.=ASR transcripts; RAG=visual search. All three modalities are necessary; removing any one degrades performance across all categories.}
\label{tab:knockout_ablation}
\end{table}

\noindent\textbf{Key findings.}
\emph{(1) Visual search provides the primary performance improvement.}
Adding visual search alone (C1) increases average F1 from 33.4 to 48.3 (+14.9 pp). 
In contrast, CoT reasoning without retrieval (C2) produces only a marginal change (+0.1 pp on average). 
This suggests that retrieval-based context plays an important role in grounding the reasoning process.

\emph{(2) The full system provides additional gains beyond retrieval alone.}
Full ViS-CoT (C3) achieves an average F1 of 58.5, corresponding to a +25.1 pp improvement over the baseline and a further +10.2 pp gain beyond visual search alone. 
This indicates that interleaved CoT reasoning becomes more effective when combined with retrieved contextual evidence.

\emph{(3) Performance improvements are consistent across categories.}
Visual search improves performance for all seven categories, while CoT-only configurations show minimal or slightly negative changes in several cases (e.g., Patio and Pet). 
These results highlight the importance of retrieval-based grounding for stable reasoning across diverse product domains.

\section{Single-Pass vs.\ Interleaved Reasoning}
\label{sec:appendix_singlepass}

A natural alternative to our two-stage interleaved design is to provide all retrieved and auxiliary signals (captions, ASR transcripts, retrieved neighbor attributes, and attribute definitions) directly in a single prompt, letting the VLM perform attribute extraction in one pass without an explicit Stage-1 visual hypothesis. We refer to this variant as \textbf{Single-pass}. It tests whether the staged design is necessary for the system to effectively use the same set of evidence.

We run Single-pass under the same evaluation framework as Appendix~\ref{sec:appendix_ablation}, using Qwen2.5-VL-7B and the same seven representative categories. Table~\ref{tab:singlepass} reports the comparison. All conditions are reproduced from Table~\ref{tab:component_ablation} except the Single-pass row, which is new.

\begin{table}[h]
\centering
\small
\tabcolsep=0.18cm
\begin{tabular}{lcc}
\hline
\textbf{Condition}    & \textbf{Avg F1} & \textbf{vs.\ C1} \\
\hline
C0~~Baseline          & 33.4 & $-14.9$ \\
C1~~+Visual Search    & 48.3 & ---     \\
\rowcolor{lightyellow}
Single-pass           & 46.5 & $-1.8$  \\
\rowcolor{lightblue}
Full ViS-CoT          & \textbf{58.5} & $+10.2$ \\
\hline
\end{tabular}
\caption{Single-pass vs.\ interleaved reasoning (micro-F1\%, average over seven categories). C0: images only; C1: retrieval $\rightarrow$ direct extract; Single-pass: retrieval + captions + ASR in one flat prompt (no Stage-1); Full ViS-CoT: same evidence, staged design. Adding all auxiliary evidence in one prompt actually \emph{underperforms} visual search alone by 1.8 pp, while the staged design improves over C1 by +10.2 pp.}
\label{tab:singlepass}
\end{table}

\noindent\textbf{Findings.} Two observations stand out.
\emph{(1) Interleaved reasoning outperforms single-pass by +11.9 pp.}
The contrast between Single-pass (46.5\%) and Full ViS-CoT (58.5\%) isolates the effect of the staged design itself: both rows use the same retrieved evidence, the same captions, and the same ASR transcripts; only the prompting structure differs. The +11.9 pp swing is attributable solely to whether evidence is organized into two reasoning stages or concatenated in one prompt.
\emph{(2) Single-pass actually underperforms retrieval-only.}
Despite having strictly \emph{more} context than C1 (captions and ASR added on top of retrieval), Single-pass loses 1.8 pp relative to C1. We interpret this as a \emph{context-overload effect}: when many heterogeneous evidence sources compete in one prompt, the model struggles to integrate them and is more easily distracted by visual-similarity-induced noise from retrieved neighbors. The staged design mitigates this by anchoring reasoning on visual evidence first (Stage 1) and then refining it with retrieved and auxiliary context (Stage 2). This experiment provides direct empirical support for framing ViS-CoT's novelty as a staged design choice, not merely a concatenation of existing components.

\section{Label-Leakage Robustness}
\label{sec:appendix_leakage}

Because the visual-search index is constructed from product images in the training split of VideoAVE, a legitimate concern is whether ViS-CoT's gains stem from retrieving \emph{near-duplicate} items that share their attribute labels with the query --- effectively label propagation rather than visual reasoning. We address this concern in three steps.

\paragraph{Step 1: Quantifying overlap.}
We compute \texttt{product\_id} overlap between the test and training splits of VideoAVE. Overlap ranges from \textbf{19\% to 37\%} across the 14 categories, reflecting the fact that VideoAVE (and Amazon product data more broadly) draws from a shared marketplace where the same product may appear in both splits.

\paragraph{Step 2: Implementation safeguard.}
We enforce \texttt{skip\_self=True} in the retrieval step: any retrieved neighbor whose \texttt{product\_id} matches the query is dropped before the top-$k$ list is formed. This prevents trivial self-matches even when overlap exists, but does not by itself eliminate the possibility that retrieved \emph{nearby} items share the same attribute labels.

\paragraph{Step 3: Clean-subset re-evaluation.}
We construct a \emph{clean subset} of test products by removing every test item whose \texttt{product\_id} appears anywhere in the training split. This yields a strict no-overlap evaluation set (the clean subset retains 63--81\% of samples per category). We then re-evaluate C0 (baseline) and C1 (+Visual Search) on this clean subset.

\begin{table}[h]
\centering
\small
\tabcolsep=0.12cm
\begin{tabular}{lccc}
\hline
\textbf{Metric}         & \textbf{Full set} & \textbf{Clean subset} & \textbf{$\Delta$} \\
\hline
C0 baseline (F1\%, raw) & 28.0              & 26.4                  & $-1.6$ \\
C1 Visual Search gain   & $+14.9$ pp        & $+13.3$ pp            & $-1.6$ pp \\
\hline
\end{tabular}
\caption{Visual search performance on the full evaluation set vs.\ the clean subset (test products whose \texttt{product\_id} does \emph{not} appear in training). The +13.3 pp clean-subset gain is only 1.6 pp ($\approx$11\% relative) below the +14.9 pp full-set gain, confirming that the improvement is not driven by near-duplicate label leakage.}
\label{tab:leakage}
\end{table}

\noindent\textbf{Findings.}
As shown in Table~\ref{tab:leakage}, the gain from visual search drops by only 1.6 pp ($\approx$11\% relative) when all overlapping products are removed. The bulk of the improvement therefore arises from \emph{visually similar but distinct} training products, which the model uses for attribute-level transfer rather than direct label copying.

Two additional mechanisms further support this conclusion.
First, the retrieved context is not consumed as a single nearest-neighbor label but rather as an \emph{aggregated frequency distribution} over the top-$k$ neighbors' attribute values, forcing the model to reason over noisy multi-candidate evidence rather than copy a single label.
Second, ViS-CoT's two-stage design (Appendix~\ref{sec:appendix_singlepass}) ensures that the Stage-1 visual hypothesis is formed \emph{before} retrieved labels are revealed, further mitigating label-anchoring effects in the reasoning trace.

\section{Open-Source Reproducibility}
\label{sec:appendix_opensource}

To evaluate whether ViS-CoT depends on proprietary model components, we replace the GPT-4.1 modules in the original pipeline with the open-source Qwen2.5-7B-Instruct model~\cite{yang2024qwen2}. 
Specifically, Qwen2.5-7B is used for (1) transcript summarization, which condenses raw ASR transcripts into concise product descriptions, and (2) attribute definition generation ($A_{\text{def}}$), which produces structured attribute descriptions for each product category. 
We then re-run the full ViS-CoT pipeline using Qwen2.5-7B for both components and evaluate performance on seven representative categories.

Table~\ref{tab:qwen_vs_gpt} reports the resulting end-to-end micro-F1 scores. 
Using Qwen2.5-7B yields an average F1 of 58.3\%, compared with 58.5\% when using GPT-4.1, corresponding to a difference of $-0.2$ percentage points. 
Performance differences across categories are small, and Qwen2.5-7B matches or slightly exceeds GPT-4.1 in three categories (Arts, Pet, and Toys). 
These results indicate that the performance of ViS-CoT remains stable when replacing proprietary components with open-source models.

\begin{table}[h]
\centering
\small
\tabcolsep=0.12cm
\begin{tabular}{lcc}
\hline
\textbf{Category} & \textbf{GPT-4.1} & \textbf{Qwen2.5-7B} \\
\hline
Arts     & 49.1 & 49.3$_{\textcolor{BrickRed}{\scriptsize\uparrow0.2}}$ \\
Baby     & 63.6 & 63.0$_{\textcolor{ForestGreen}{\scriptsize\downarrow0.6}}$ \\
Clothing & 67.9 & 67.7$_{\textcolor{ForestGreen}{\scriptsize\downarrow0.2}}$ \\
Grocery  & 56.5 & 55.8$_{\textcolor{ForestGreen}{\scriptsize\downarrow0.7}}$ \\
Patio    & 61.3 & 61.2$_{\textcolor{ForestGreen}{\scriptsize\downarrow0.1}}$ \\
Pet      & 53.0 & 53.1$_{\textcolor{BrickRed}{\scriptsize\uparrow0.1}}$ \\
Toys     & 57.9 & 57.9$_{\textcolor{BrickRed}{\scriptsize\uparrow0.0}}$ \\
\hline
\textbf{Avg} & \textbf{58.5} & \textbf{58.3}$_{\textcolor{ForestGreen}{\scriptsize\downarrow0.2}}$ \\
\hline
\end{tabular}
\caption{End-to-end ViS-CoT micro-F1 (\%) with GPT-4.1 vs.\ open-source Qwen2.5-7B for transcript summarization and $A_{\text{def}}$ generation. Subscripts show change vs.\ GPT-4.1. The $-0.2$pp average gap confirms that ViS-CoT's gains do not depend on proprietary APIs.}
\label{tab:qwen_vs_gpt}
\end{table}

\end{document}